# Parallel Texts in the Hebrew Bible, New Methods and Visualizations

Martijn Naaijer[1]*, Dirk Roorda[2]

[1]Vrije Universiteit Amsterdam, The Netherlands

[2]Data Archiving and Networked Services, The Netherlands

*Corresponding author: m.naaijer@vu.nl

**Abstract**

In this article we develop an algorithm to detect parallel texts in the Masoretic Text of the Hebrew Bible. The results are presented online and chapters in the Hebrew Bible containing parallel passages can be inspected synoptically. Differences between parallel passages are highlighted. In a similar way the MT of Isaiah is presented synoptically with 1QIsa[a]. We also investigate how one can investigate the degree of similarity between parallel passages with the help of a case study of 2 Kings 19-25 and its parallels in Isaiah, Jeremiah and 2 Chronicles.

**keywords**

Hebrew Bible, parallel texts, linguistic data, visualizations

## 1. INTRODUCTION

The reuse of texts is a well-known phenomenon in the Hebrew Bible. For example, in the book of Genesis we find various allusions to older Mesopotamian texts in the stories in the story of the flood [Hallo, 2003: 458-460]. There are also many cases of inner biblical intertextuality, such as the references to the story of Jacob and Esau in Hosea 1 and of course there exist numerous retellings of the Hebrew Bible in later ancient literature, such as the Genesis Apocryphon and the book of Jubilees, also called "Rewritten Scripture" [Parry and Tov, 2004]. In this article we focus on parallel texts within the Hebrew Bible. There exists a lot of literature on this issue, but there is no open source online tool available yet with which it is possible to search for parallel texts and with which it is possible to present these parallels synoptically and online[1]. There is no fixed definition of what exactly parallel texts are. [Endres, Millar and Burns, 1998: ix] describes them as:

---

[1] There are various printed synopses. For a comparison of several of these, see [Verheij, 1992].

"…more than one "text"—whether clauses, sentences, verses or longer sections—manifest similar language, tone or structure. Parallels may be verbatim, nearly verbatim, structural, within the same book of the Bible, between different books of the Bible…"

In this research we will only study parallels with a certain minimal similarity of lexical characteristics. We developed a tool with which it is possible to find parallel texts within the Hebrew Bible and to visualize the results synoptically. In a similar we visualize the book of Isaiah in the MT and 1QIsa$^a$.

In section 2 the relevance of the study of parallel passages in the study of linguistic variation in Biblical Hebrew is described. The data, the software and the tool are described in section 3. The tool consists of two Jupyter notebooks, which can be found online[2]. They form an inseparable part of this article. In section 4 we take a closer look at a specific group of parallel texts, namely 2 Kings 19-25 and its parallels in the books of Isaiah, Jeremiah and 2 Chronicles and we present a method by which it is possible to compare different manuscripts of the book of Isaiah.

## 2. PARALLEL PASSAGES AND LINGUISTIC VARIATION IN BIBLICAL HEBREW

Parallel passages in the Hebrew Bible play an important role in reconstructing the development and growth of biblical texts and also in the study of the linguistic variation in the Hebrew language they are used often to explain diachronic change.

In most studies on diachrony in Biblical Hebrew, Archaic Biblical Hebrew, Early Biblical Hebrew and Late Biblical Hebrew are distinguished[3]. According to this chronological model Archaic Biblical Hebrew can be found in several poems in the Hebrew Bible [Sáenz-Badillos, 1993: 56-62] [Notarius, 2013]; the books of the Pentateuch and the Former Prophets[4] are the best examples of Early Biblical Hebrew and the so called core late books[5] are written in Late Biblical Hebrew. By comparing a characteristic late linguistic feature in one of the late books with its corresponding early alternative, scholars are able to uncover the chronological development of

---

[2] https://shebanq.ancient-data.org/tools?goto=parallel
[3] For instance "A History of the Hebrew Language" by [Sáenz-Badillos, 1993] is based on this division.
[4] The Pentateuch consists of the books of Genesis, Exodus, Leviticus, Numbers and Deuteronomy, the Former Prophets consist of the books of Joshua, Judges, Samuel and Kings.
[5] The books of Esther, Daniel, Ezra, Nehemiah and Chronicles are the core late books. Sometimes other books, like Qoheleth is also considered to be late by many, but not everyone is certain, see for instance [Hurvitz, 2014: 4] and [Young, 1993: 140-157]

Biblical Hebrew. Very often scholars make use of examples in parallel texts, especially in the early books of Samuel and Kings and their late parallels in Chronicles to show this development. [Kropat, 1909] is a classic study of the syntax of the book of Chronicles in which this approach is followed consequently.

A well-known example is the use of the prepositions אל and על. At various places in the Hebrew Bible we find אל in the early books Samuel and Kings, whereas על can be found in the parallel late texts in Chronicles. For example:

1 Samuel 31:3 וַתִּכְבַּד הַמִּלְחָמָה אֶל־שָׁאוּל

1 Chronicles 10:3 וַתִּכְבַּד הַמִּלְחָמָה עַל־שָׁאוּל

2 Kings 22:16 הִנְנִי מֵבִיא רָעָה אֶל־הַמָּקוֹם הַזֶּה

2 Chronicles 34:24 הִנְנִי מֵבִיא רָעָה עַל־הַמָּקוֹם הַזֶּה

[Rooker, 1990: 127] gives a long list of examples of this shift, and indicates that in parallels between the books of Kings and Chronicles, the reversed pattern can be found three times:

1 Kings 15:20 אֲשֶׁר־לוֹ עַל־עָרֵי יִשְׂרָאֵל

2 Chronicles 16:4 אֲשֶׁר־לוֹ אֶל־עָרֵי יִשְׂרָאֵל6

Another phenomenon that can be observed in parallel texts is the more abundant use of matres lectionis in Late Biblical Hebrew. We can see this difference often in parallels containing the name David [Hurvitz, 2014: 88-89]:

2 Samuel 5:17 דָּוִד

1 Chronicles 14:8 דָּוִיד

In many other words, Chronicles has an increased number of matres lectionis, compared to the books of Samuel and Kings, for instance in 1 Chronicles 17:8 we find four extra cases compared to its parallel 2 Samuel 7:9[7].

---

[6] The other cases are 2 Kings22:16 vs. 2 Chronicles 34:15 and 2 Kings 22:20.
[7] See https://shebanq.ancient-data.org/shebanq/static/docs/tools/parallel/files/chapters/Samuel_II_7_vs_Chronica_I_17.html

The model in which an important part of the variation in Biblical Hebrew is described in terms of a chronological change and the related method of linguistic dating have been contested in recent years, see especially [Young, Rezetko and Ehrensvärd, 2008] and [Rezetko and Young, 2014]. It is beyond the scope of this article to discuss the whole issue, but a point of critique in relation to the use of parallel texts is that generally only those examples of parallels are shown that confirm the presupposed theory of diachronic variation in Biblical Hebrew.

## 3. THE TOOL

### 3.1 Open source tools

There is an important requirement that accompanies the use of digital methods in scholarly research: they should not diminish the transparency of the ways that hypotheses are confirmed or rejected. Moreover, computations that lead to new results should be replicable by other scholars. For this reason, it is as essential to publish the data and digital tools as it is important to publish the articles in which the conclusions are stated and discussed. Replication of results obtained by software is in general not easy, because the digital world is in constant flux and all software goes from version to version. To lower the barrier for effective replication we take care that our data is properly archived and our software is available as Open Source in online repositories with versioning. These resources can then be referenced in a persistent way, e.g. through Digital Object Identifiers (DOIs) and can be freely downloaded.

### 3.2 Database, website and tools

We base our reasoning on the database of the Eep Talstra Centre for Bible and Computer (ETCBC). A few years ago this database has been brought fully online as a research tool in the form of SHEBANQ (System for HEBrew text: Annotations for Queries and markup)[8]. On this website one can read the complete Masoretic Text (abbreviated as MT) according to the Biblia Hebraica Stuttgartensia[9]. The linguistic information of the ETCBC database is exposed and users can perform queries on the combined information of text and linguistic data. The underlying format is Linguistic Annotation Framework (LAF)[10], and we have built a tool, LAF-Fabric[11], to perform data analysis on the Hebrew Bible as represented in LAF.

---

[8] https://shebanq.ancient-data.org
[9] This text is also the basis of our analyses in the notebooks.
[10] ISO standard 24612:2012. Edition 1, 2012-06-15, http://www.iso.org/iso/home/store/catalogue_tc/catalogue_detail.htm?csnumber=37326

We refer to [Van Peursen et al., 2015] for the data, and to [Roorda, 2015a] for the SHEBANQ software and to [Roorda, 2015b] for LAF-Fabric and supporting tools[12]. The history of this Hebrew Text database is told by [Roorda, 2015] and more about the underlying models for processing text and data can be found in [Roorda and Van Peursen, 2016].

### 3.3. Chasing parallel passages

We use an algorithm to find parallel passages. This algorithm is implemented and documented in an executable notebook called Parallels, which is available from SHEBANQ[13].

Finding parallel passages is not easy. In the first place, the concept of a parallel passage has no clear-cut definition. Secondly, the notion of correspondence between parallel passages may involve characteristics that are difficult to compute from the text, e.g. semantics. And, thirdly, we have to tune our algorithm to the subtle boundary between too much and too little similarity. In the Parallels notebook we have chosen to base similarity on lexical characteristics only. We have defined two different ways of measuring such similarity, and used them both in a number of experiments (165 in total). We have used automatic indicators to evaluate the result of those experiments, which yielded 18 good outcomes. The present article is based on the parallels found by those "good" experiments.

The first step of our experiments consisted of chunking the Hebrew Bible. We made chunks of a fixed size (10, 20 50 or 100 words) and chunks related to objects in the database. In the latter case, each chunk consists of a sentence, a half-verse, a verse, or a complete chapter.

After splitting the Hebrew Bible in chunks each chunk is compared with each other chunk, to be able to detect the similarity between these chunks. This was done in two ways. The first similarity method (called SET) works as follows. We take from both passages the set of lexemes. So we ignore order, multiplicity of occurrences, and the shapes of the concrete textual occurrences. We count the number of lexemes in the intersection and divide it by the number of lexemes in the union. If both passages use exactly the same lexeme set, the similarity is 1, and if they are based on totally different lexemes, their similarity is 0. If we require a significant level of SET similarity, (60 or more) we find indeed most of the known similarities. The second similarity method (called LCS) is based on the Levenshtein distance of the two passages. This is a measure of how much editing is needed to transform one string into the other[14]. Finally, we searched for cliques, these are chunks with a similarity score above a certain threshold. We varied this threshold between 30% and 100%. The results of all these experiments can be found in the table of section 1. Results of the notebook. If we set our similarity threshold to high, we miss parallels. If we set it to low, we tend to get gigantic, drifting cliques, with unrelated passages connected by chains of "similar" passages. We compare the outcomes for different choices of the similarity threshold, and we have an objective criterion for finding the sweet spot: the ratio between the number of cliques and the length of the longest clique.

---

[11] http://laf-fabric.readthedocs.org
[12] An overview of all sources is available at https://shebanq.ancient-data.org/sources .
[13] https://shebanq.ancient-data.org/tools?goto=parallel

In a second notebook, kings_ii[15] we apply the techniques of finding parallel texts to our reference chapters, 2 Kings 19-25. We have used the SET method to make an inventory of all verses that are potentially parallel to our reference chapters in the books of Isaiah, Jeremiah and 2 Chronicles. After that, we have applied the LCS method on all pairs of sentences in those verses, in order to study patterns in the differences between all those parallel verses.

## 4. RESULTS

In this section we discuss the results of our experiments, which are described in the notebooks. First the results in the Parallels notebooks are discussed, in which parallels in the complete Masoretic text are detected, then we describe the parallels of 2 Kings 19-25.

### 4.1. The Parallels Notebook

The table in section 1. Results of the Parallels notebook shows the results of the experiments done. This table shows that using a fixed chunk size of a certain number of words did not lead to promising results. The best results were obtained by comparing chunks related to specific objects in the database (green areas can be found in the case of chunking in half-verses, verses, and chapters) and with a threshold that is not too low (in that case too many cliques are found) and that is not too high (then some good cases of parallel texts might be overlooked). These best results are colored green and in these cases we made binary chapter comparisons, which can be accessed by clicking on the hyperlink in the cells. In the chapter comparison one can find the complete chapters containing at least one clique with a colorful presentation of the similarities and differences of the verses in these chapters. For instance, in the experiment using verse chunks, method SET and threshold 85, we find the pair of chapters 2 Samuel 22 and Psalm 18[16]. This Song of David is a well-known parallel in the Hebrew Bible and in our presentation one can easily see where the similarities and differences are. In the chapter comparison identical parts are white, green and red indicates a plus in the text and yellow indicates a substitution. These pluses and substitutions are only a formal, computational indication, they do not mean that a certain scribe or editor added or changed parts of the text.

---

[14] A precise, mathematical description of the similarity methods can be found in the Parallels notebook, section 3. Method description.
[15] See also https://shebanq.ancient-data.org/tools?goto=parallel

In the index of the binary chapter comparison in the same experiment one can find most of the well-known parallels in the Hebrew Bible. It finds the parallels between Samuel, Kings and Chronicles, between Ezra 2 and Nehemiah 7 and parallel texts in the Psalms, for instance Psalm 60 versus Psalm 108. It also finds more formulaic parallel verses, like Job 34:1/Job 35:1 and Job 38:1 and 3/Job 40:6-7.

On the other hand, there exist also a number of parallels that are not detected by our tool. The parallel between 1 Chronicles 1:1-4 and Genesis 5:1-32 is not found and it is also directly clear why. In Chronicles these verses consist only of a list of proper names, whereas Genesis tells a short biography of each of these persons [Endres, Millar and Burns, 1998: 3-4] [Bendavid, 1972: 14-15]. This parallel is based on the order of the persons, instead of the lexical similarities in our chunks. Another example with which the tool has difficulties is 2 Kings 18:13-19:37 and Isaiah 36:1-22 on the one hand and 2 Chronicles 32: 9-23 on the other [Endres, Millar and Burns, 1998: 307-316] [Van Peursen and Talstra, 2007: 52:71]. It is clear that these parallels deal with the same events, Sennacherib's campaign, but the parallel texts strongly vary in their wording of the story. In some of our experiments the tool finds cliques[17], but in the corresponding chapter comparison we do not find these similarities in a clear way[18].

Although there many overlapping results between the different experiments, there are also differences, so what is considered a parallel in one experiment is excluded in another experiment. This leads to the question what exactly is a parallel. The answer is that that depends completely on the needs and the questions of the researcher. One can imagine that for some kind of research one is interested in parallels that consist of two specific words occurring together. These parallels are generally excluded in our experiments, but it would be easy to write such a program with the use of LAF-Fabric. On the other hand, others might be interested in parallels that stretch over more verses. For that kind of research our tool generally works very well.

---

[16] It can be accessed here: https://shebanq.ancient-data.org/shebanq/static/docs/tools/parallel/files/chapters/Samuel_II_22_vs_Psalmi_18.html

[17] For instance in clique 1443 in: https://shebanq.ancient-data.org/shebanq/static/docs/tools/parallel/files/experiments/O_half_verse_SET_M50_S80/clique_O_half_verse_SET_M50_S80_28.html#c_1443

[18] https://shebanq.ancient-data.org/shebanq/static/docs/tools/parallel/files/chapters/Reges_II_18_vs_Chronica_II_32.html

**4.2. A case study: 2 Kings 19-25**

Now we have a closer look at the final part of the book of Kings and their parallel texts. This is a group of interesting chapters, because 2 Kings 19-20 have parallels in Isaiah 37-39, 2 Kings 21-23 have parallels in 2 Chronicles 33-34 and 2 Kings 24-25 have parallels in Jeremiah 52, as can be seen in the chapter comparisons of these books[19].

As in the previous section, we focus mainly on the formal characteristics of textual similarities.

*4.2.1. Visualization of 2 Kings and its parallels*

First we have found out where all the parallel verses of 2 Kings 19-25 in the MT can be found and graphed the results. The similarity between parallel passages was based on verses and the similarity was based on the SET method, and a similarity higher than 60 is considered parallel. The results are graphed in figure 1, Parallels involving 2 Kings 19-25 in the notebook[20]. The backbone of this figure is formed by Reges_IIr, and a grey line from this column to one of the other columns indicates a parallel passage. Passages with high similarities are linked with darker lines than passages with lower similarity. In the figure we find the parallel verses in Isaiah, 2 Chronicles and Jeremiah, but some other passages are found in the books of Exodus, Deuteronomy, Ezechiel, Haggai, and elsewhere in the book of Kings. Also provided is a complete synoptic representation of all the parallel verses[21]. By playing with the similarity threshold in the program one can include or exclude passages.

In the synoptic representation of the parallel verses one can see that many of the parallels are of a more or less formulaic nature, for instance in the case of Ezekiel 12:8, Jeremiah 37:6, and

---

[19] For printed parallels of these texts in Hebrew with commentary, see [Person, 1997].
[20] https://shebanq.ancient-data.org/shebanq/static/docs/tools/parallel/kings_parallels.pdf
[21] https://shebanq.ancient-data.org/shebanq/static/docs/tools/parallel/kings_parallels_h.html

Haggai 1:3[22] and most of the parallels within the book of Kings[23]. There are also some parallels that are often not considered real parallels, for instance Deuteronomy 8:6 and Jeremiah 2:17, both parallel with 2 Kings 21:22. These examples show that the notion of what is parallel strongly depends on ones definition of parallel passages.

*4.2.2. The similarity of parallel passages*

How similar are similar passages and how can various parallels be compared with each other quantitatively? In this section we would like to find out which of the parallels of 2 Kings 19-25 in the books of Isaiah, Jeremiah, Kings and Chronicles is closest to the text in the book of Kings. We do not presuppose that Kings is the earlier or original text, the focus lies exclusively on measuring the similarities.

First all sentences were extracted in the books of Isaiah, Jeremiah and 2 Chronicles that have a parallel in 2 Kings 19-25. Then a file was made in which the similarity of all sentences in 2 Kings 19-25, Isaiah 37-39, 2 Chronicles 33-34, and Jeremiah 52 are compared with each other[24]. Most sentences that are compared with each other in this file are not similar, but some are. We want to find out which of the groups of parallel passages are most similar.

Figure 1 shows a boxplot of the similarities of all the relevant sentences in 2 Kings 19-25 on the one hand and the relevant sentences in Isaiah, Jeremiah and 2 Chronicles together on the other hand.

---

[22] These verses are parallel with 2 Kings 21:10.
[23] For example 1 Kings 14:29, 15:7, 15:31, 16:5, 16:14, 16:27, 22:46, 2 Kings 8:23, 10:34, 12:20, 13:8, 13:12, 14:8, 15:6, 15:21, 15:26, 15:31, 15:36, 16:19 are all parallel with 2 Kings 21:17 and 1 Kings 15:11, 2 Chronicles 26:4, 29:2, 33:22, 2 Kings 14:3, 2 Kings 15:3, 15:34, 18:3, and Jeremiah 52:2 are all parallel with 2 Kings 23:32.
[24] The file can be downloaded here: https://shebanq.ancient-data.org/shebanq/static/docs/tools/parallel/kings_similarities.tsv. The first four columns in this file define the first sentence in the comparison (book_1, chapter_1, verse_1, num_1), the variable num_1 stands for number of the sentence in the verse, the second four variables define the second sentence (book_2, chapter_2, verse_2, num_2). The last column indicates the similarity between the two sentences.

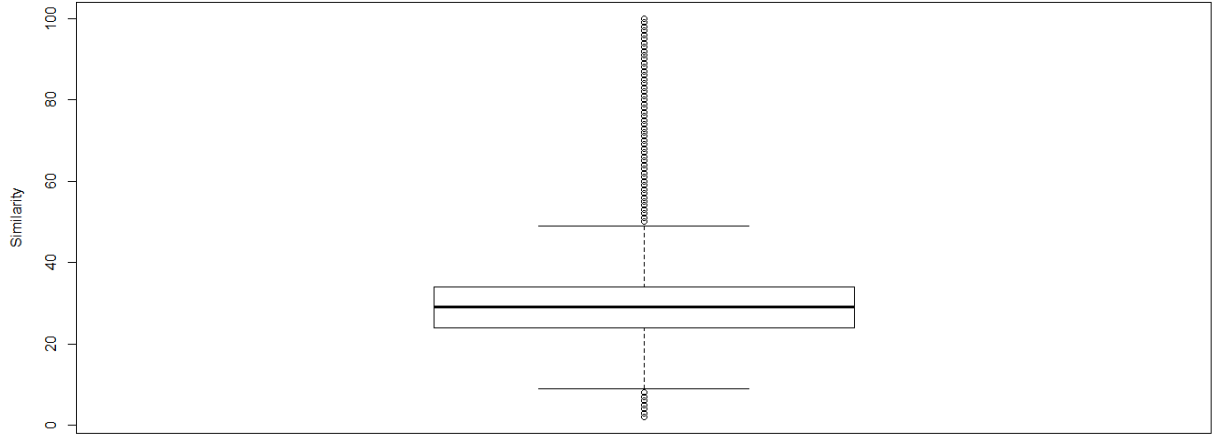

Figure 1. Sentence similarities between 2 Kings 19-25 and Isaiah, Jeremiah and 2 Chronicles together.

The median of all the values is 29 and the mean is with 29.26 pretty close to the median value. The violin plot in figure 2 shows the corresponding densities.

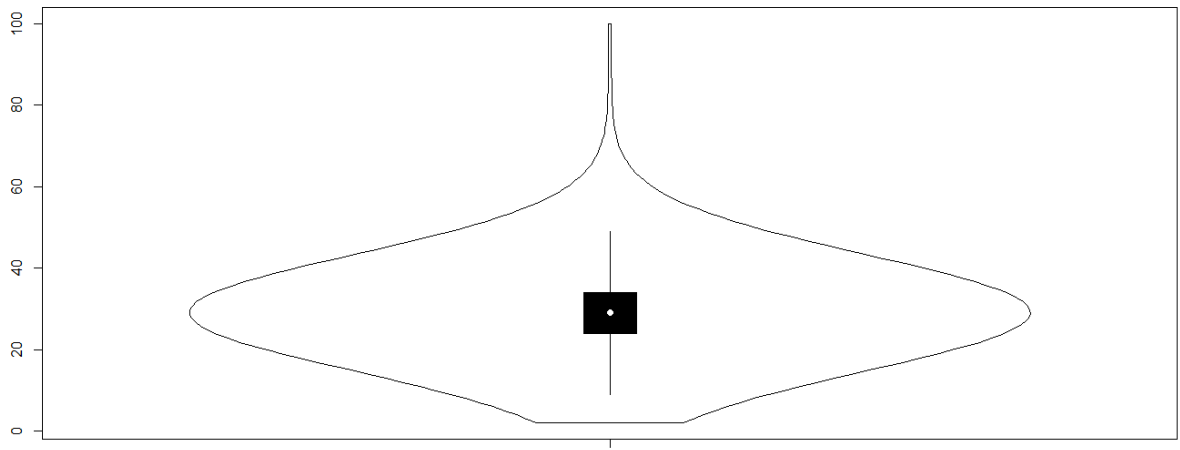

Figure 2. Density of sentence similarities between 2 Kings 19-25 and Isaiah, Jeremiah and 2 Chronicles together.

There is a very high density around the median value, which rapidly diminishes when higher or lower values are approached. The parallel sentences can be found in the region with a similarity higher than 60. In the violin plot one can see that this is only a small minority of the sentences. Figure 3 shows the results if the books are split, the situation is more or less identical to that of the previous boxplot, which indicates that the general distribution for each of the books are more or less identical.

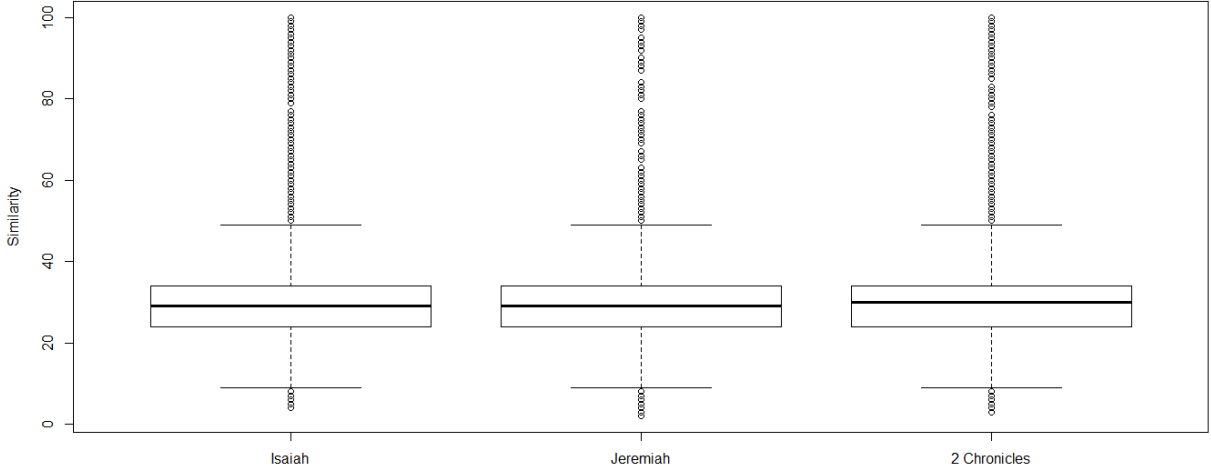

Figure 3. Sentence similarities between 2 Kings 19-25 and Isaiah, Jeremiah and 2 Chronicles separated.

The situation changes if the large bulk of data around the median is neglected and we select only those sentences with a similarity higher than 60. This is the region in which the parallel passages are found, and figure 4 is a boxplot of the books involved.

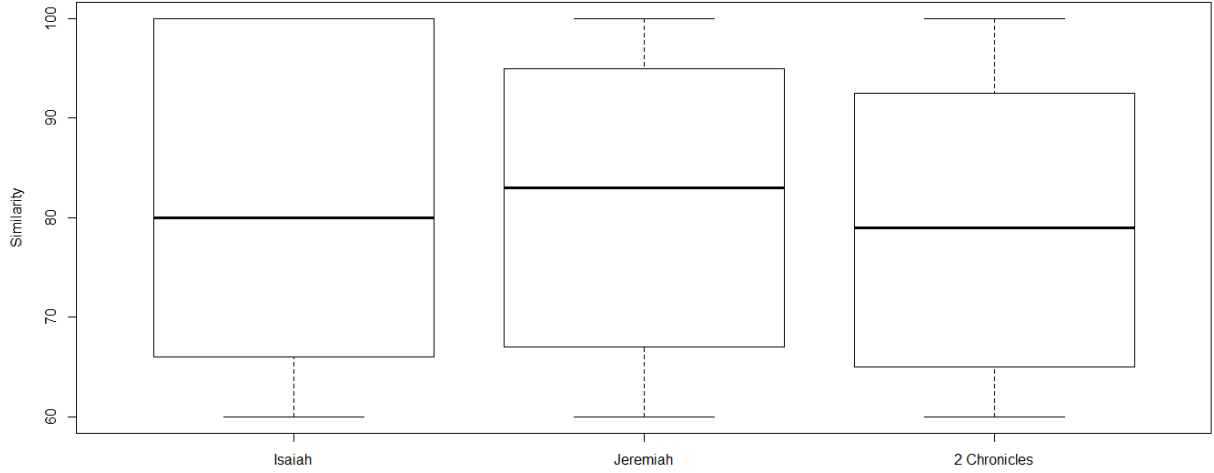

Figure 4. Similarity between similar sentences in 2 Kings 19-25 and Isaiah, Jeremiah and 2 Chronicles (similarity >= 60).

The medians of all three books are around 80, these do not differ very much. The upper part of the box of Isaiah is higher than that of the other books, which is an indication of a higher amount of very similar sentences. This becomes clearer if only those sentences are selected with a similarity higher than 80. Figure 5 shows the boxplot of this comparison.

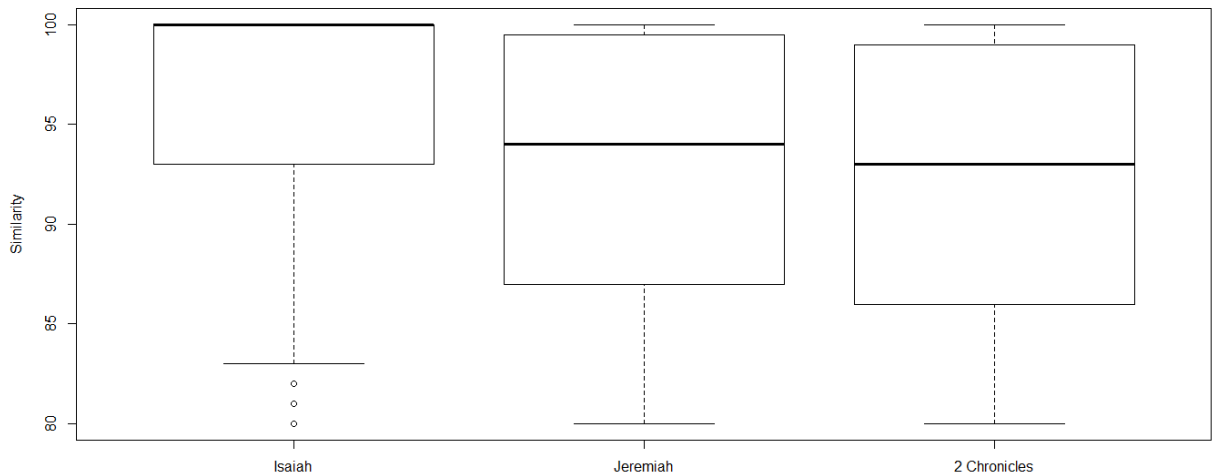

Figure 5. Similarity between similar sentences in 2 Kings 19-25 and Isaiah, Jeremiah and 2

Chronicles (similarity >= 80).

It shows that the similar sentences between Isaiah and 2 Kings 19-25 are far more similar to each other than the parallels between 2 Kings 19-25, Jeremiah and 2 Chronicles and this difference is significant[25]. The boxplots show the distribution of the similar values, but it is also striking to see that the number of verses with a similarity higher than 80, is far higher in Isaiah than in Jeremiah and 2 Chronicles[26].

We can conclude that there is a kind of base level of similarity between the biblical passages under consideration of 29, which is the median value. All books show more or less the same pattern around this median. It is only in the area of very high similarity (>= 80) that Isaiah is more similar its parallels in 2 Kings than the parallels in Jeremiah and 2 Chronicles. We can conclude that Isaiah has a relatively long parallel with 2 Kings and those sentences that are parallel are also very similar to their counterparts in 2 Kings[27].

*4.2.3. Different manuscripts of Isaiah*

There is an important role in biblical scholarship for the study of the biblical text in different manuscripts. The parallels in the book of Kings and Isaiah as discussed in the previous section can all be found in the MT, but the biblical manuscripts of the Dead Sea Scrolls offer an enormous wealth of material with which the MT can be compared. The manuscript known as the Great Isaiah Scroll or 1QIsa[a] contains more or less the complete text of the book of Isaiah[28]. We have created a binary chapter comparison for the complete text of the text of Isaiah in the MT and 1QIsa[a]. The result can be found in section 0.1 Results of the notebook[29].

## 5. CONCLUSIONS

With the use of some simple methods we have explored how parallel texts in the Hebrew Bible can be detected in a and we developed a way in which parallel passages can be studied synoptically online. With this approach we found most of the parallels that can be found in printed overviews of parallel texts. The parallel texts that were not found are those that are not based on lexical similarities, such as 1 Chronicles 1:1-4 and Genesis 5:1:32. An advantage of our approach is that it is completely reproducible. This gives other researchers also the possibility to adapt the experiments and synoptic representations in a way that fits their needs, which is an advantage over printed editions, in which always a certain choice has to be made for a specific kind of representation of the texts.

---

[25] The t-tests of Isaiah versus Jeremiah and Chronicles resulted both times in p < 0.001. Jeremiah versus 2 Chronicles resulted in p = 0.5. We applied a Bonferroni correction (alpha = 0.05/3).

[26] Of the group of sentences with a similarity higher than 80, Isaiah contains 182 sentences, Jeremiah contains 68 sentences, and 2 Chronicles contains 84 sentences.

[27] Although on other grounds, this is also the opinion of [Van Peursen and Talstra, 2007].

[28] For printed texts of the scroll, see [Parry and Qimron, 1998], [Ulrich, 2010]. For a full digital facsimile, see http://dss.collections.imj.org.il/isaiah#. Here one can also find a parallel translation of the MT and the Great Isaiah Scroll: http://dss.collections.imj.org.il/chapters_pg. For the text with variants and commentary, see [Ulrich, Flint and Abegg, 2010]. An important study of the language of 1QIsa[a] is [Kutscher 1974].

[29] See: https://shebanq.ancient-data.org/shebanq/static/docs/tools/parallel/Isaiah-mt-1QIsaa.html

More important than the visual representation of the parallels are the underlying data. If these are not generally accessible it is difficult to check how the data, the methods and the visualizations are related. Only if the data are open source this goal can be reached and in science this is the only scientific way to proceed.

The work done in this project is not finished yet, in the sense that there are no improvements possible. We made visual representations of a few chapters of MT Isaiah and 1QIsa$^a$, but it would be good if all Dead Sea Scrolls could be publicly available and easily comparable with the MT. Another issue is how more than two texts or manuscripts can be compared in an efficient way.

In many traditional studies on textual development of the Hebrew Bible or diachrony in Biblical Hebrew very often the results are driven by presuppositions on how various texts are related to each other and how the Hebrew language has developed through time and how the text of the Hebrew Bible has been transmitted until the moment the manuscripts that we possess were created. With our research we would like to take a step back from these ideas and presuppositions to be able to study the data freshly in a way that is accessible to everyone. We hope that this research may be a stimulus for data driven research of the Hebrew Bible.